\documentclass{ieeeaccess}
\usepackage{amsmath,amssymb,amsfonts}
\usepackage{algorithm}           % Keep this for the algorithm environment
\usepackage{algpseudocode}        % Keep this for algorithmic pseudocode
\usepackage{graphicx}
\usepackage[]{footmisc}
\usepackage{textcomp}
\usepackage[numbers,sort&compress]{natbib}

\def\BibTeX{{\rm B\kern-.05em{\sc i\kern-.025em b}\kern-.08em
    T\kern-.1667em\lower.7ex\hbox{E}\kern-.125emX}}

\begin{document}
\history{Date of publication xxxx 00, 0000, date of current version xxxx 00, 0000.}
\doi{}

\title{Multi-modal biometric authentication: Leveraging shared layer architectures for enhanced security}
\author{\uppercase{Vatchala S} \authorrefmark{1},\uppercase{Yogesh C}\authorrefmark{2}, \uppercase{Yeshwanth Govindarajan}\authorrefmark{3}, \uppercase{Krithik Raja M} \authorrefmark{3}, \uppercase{Vishal Pranav Amirtha Ganesan} \authorrefmark{3}, \uppercase{Aashish Vinod A} \authorrefmark{3} and \uppercase{Dharun Ramesh} \authorrefmark{3}}

\address[1]{Assistant professor, School of computer science and engineering, Vellore Institute of technology, Chennai-600127,India; (email:vatchala.s@vit.ac.in)}

\address[2]{Associate professor, School of computer science and engineering, Vellore Institute of technology, Chennai-600127,India; (email:yogesh.c@vit.ac.in)}

\address[3]{School of Computer Science and Engineering, Vellore Institute of Technology, Chennai 600127, India; (e-mail: yeshwanth.g2021@vitstudent.ac.in, krithikraja.m2021@vitstudent.ac.in,  vishalpranav.ag2021@vitstudent.ac.in, aashishvinod.a2021@vitstudent.ac.in, dharun.ramesh2021@vitstudent.ac.in)}

\markboth
{Author \headeretal: Preparation of Papers for IEEE TRANSACTIONS and JOURNALS}
{Author \headeretal: Preparation of Papers for IEEE TRANSACTIONS and JOURNALS}

\corresp{Corresponding author: Vatchala S (e-mail: vatchala.s@vit.ac.in).}

\corresp{This work is supported by the Vellore Institute of Technology, Chennai, India.}

\begin{abstract}
In this study, we introduce a novel multi-modal biometric authentication system that integrates facial, vocal, and signature data to enhance security measures. Utilizing a combination of Convolutional Neural Networks (CNNs) and Recurrent Neural Networks (RNNs), our model architecture uniquely incorporates dual shared layers alongside modality-specific enhancements for comprehensive feature extraction. The system undergoes rigorous training with a joint loss function, optimizing for accuracy across diverse biometric inputs. Feature-level fusion via Principal Component Analysis (PCA) and classification through Gradient Boosting Machines (GBM) further refine the authentication process. Our approach demonstrates significant improvements in authentication accuracy and robustness, paving the way for advanced secure identity verification solutions.
\end{abstract}

\begin{keywords}
Multi-modal, Convolutional Neural Networks, Recurrent Neural Networks, Principal Component Analysis, Gradient Boosting Machines
\end{keywords}

\maketitle

\titlepgskip=-21pt

\section{Introduction}
\IEEEPARstart{I} {\MakeLowercase{n}} the advent of an increasingly digital and interconnected world, the security of personal and organizational data has risen to the forefront of technological innovation and research. Within this context, biometric authentication systems, which leverage unique physiological and behavioral characteristics for identity verification, have emerged as a critical component in enhancing security infrastructures. Despite their growing ubiquity and importance, traditional biometric systems often grapple with challenges related to accuracy, privacy, and susceptibility to sophisticated attacks. To address these challenges, advancements in machine learning (ML) models have provided promising avenues for developing more secure, accurate, and efficient biometric authentication methods.
    
    This paper introduces a novel approach to biometric authentication by proposing a multi-modal system that integrates facial images, voice recordings, and signature data, leveraging the inherent strengths of each modality. Our research is driven by the hypothesis that a fusion of multiple biometric modalities, processed through a carefully architected network of shared and modality-specific layers, can significantly enhance the robustness and reliability of authentication systems. The shared layers, utilizing Convolutional Neural Networks (CNNs) and Recurrent Neural Networks (RNNs), are designed to extract and learn from the common features across modalities, while the modality-specific layers focus on the unique attributes of each biometric trait. This dual approach not only broadens the system's ability to accommodate a wider variance in biometric data but also reduces the bias inherent in single-modality systems.
    
    However, integrating diverse biometric data introduces complexities in ensuring data privacy and security, necessitating advanced solutions to protect against potential vulnerabilities. To this end, our model incorporates sophisticated data preprocessing and feature integration techniques, including Principal Component Analysis (PCA) for dimensionality reduction and Gradient Boosting Machines (GBM) for classification, enhancing the system's capability to thwart security threats effectively.

Figure 1 offers a visual representation of our innovative architecture, emphasizing a dual-layer strategy that harnesses both shared and modality-specific layers to process and analyze facial images, voice recordings, and signature data. The system's shared layers employ CNNs to analyze spatial attributes and RNNs to manage temporal data, facilitating comprehensive feature extraction across modalities. This structure allows the system to capture a rich set of biometric features, enhancing its capacity to recognize and authenticate identities with high accuracy. Following the shared processing, modality-specific layers are dedicated to refining these features further, with tailored CNNs focusing on detailed aspects of facial and signature recognition, and specific RNN configurations handling the dynamic elements of voice and signature sequences. This tailored approach ensures that each biometric characteristic's unique properties are effectively captured and integrated, reinforcing the system's robustness against various forms of spoofing and falsification. Furthermore, our sophisticated preprocessing routines, including PCA for reducing dimensionality and GBM classifiers for effective threat detection and response, equip the system with the necessary tools to offer enhanced security and privacy assurances, preparing it to handle the complexities of modern biometric authentication challenges effectively.

The primary contributions of our research are threefold and pivot around the novel shared layer mechanism. 
    \begin{itemize}
        \item Firstly, the introduction of this dual shared layer architecture represents a significant advancement in the field of biometric authentication, enabling a more sophisticated and nuanced analysis of multi-modal biometric data. 
        \item Secondly, the integration of modality-specific layers alongside the shared layers enhances the system's ability to accurately identify individuals by accommodating a broader spectrum of variability and reducing bias. 
        \item Thirdly, the fusion of learning from both the shared and modality-specific layers into a cohesive authentication decision mechanism marks a pioneering step towards achieving higher accuracy and security in biometric authentication systems.
    \end{itemize}

    \begin{figure*}
    \centering
    \includegraphics[width=1.0\textwidth]{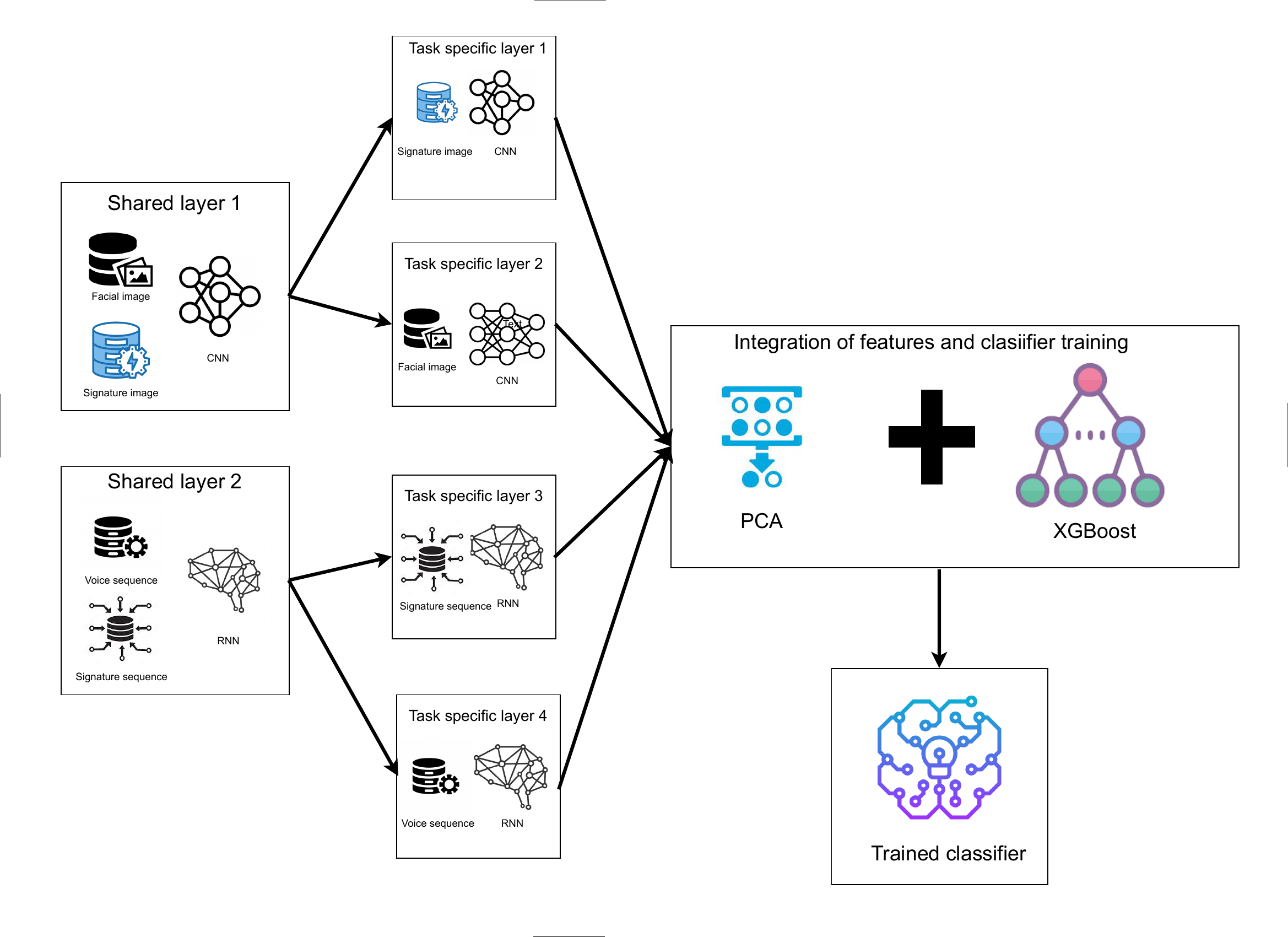}
    \caption{Architectural Diagram}
    \label{Architectural Diagram}
    \end{figure*}

 We organize the paper as follows. Section II discusses the related work and motivating factors for our proposed method. Section III outlines the methodology, including data collection and preprocessing, design of the dual shared layer architecture with modality-specific enhancements, and the training process integrating both shared and modality-specific layers. Section IV provides experimental results and performance comparisons with legacy systems, demonstrating the effectiveness of our approach. Lastly, Section V concludes with insights into the future work inspired by the outcomes of the current research.

\section{Related Works}\label{sec2}

In the burgeoning field of multimodal artificial intelligence (AI), a diverse array of applications is being explored, each harnessing the synergistic potential of integrating multiple data modalities to solve complex problems across various domains. Starting with healthcare, the Semantic Platform for Adverse Childhood Experiences Surveillance (SPACES) exemplifies the use of multimodal AI in public health, leveraging an explainable AI platform for active surveillance, diagnosis, and management of Adverse Childhood Experiences (ACEs) \cite{1}. Similarly, the rapid diagnosis and surgical prediction of necrotizing enterocolitis (NEC) demonstrate how a multimodal AI system, integrating clinical data and imaging, matches the diagnostic prowess of experienced clinicians \cite{6}.

In the domain of security and trustworthiness, multimodal fusion has been a key focus for enhancing the performance of systems in safety-critical applications. Early efforts in multimodal systems like Multimodal Dynamics \cite{2} have explored the integration of different data types for robust classification. Similarly, in fields like social media propaganda detection, multimodal AI has been successfully applied to identify patterns across diverse modalities \cite{3}. These works have paved the way for multimodal biometric systems that fuse physiological and behavioral data to achieve higher accuracy and reliability. Our approach builds on this foundation by integrating three key biometric modalities—face, voice, and signature—into a unified system, leveraging shared-layer architectures for efficient feature extraction and fusion.

Technological advances in clinical tools are also evident. CoRe-Sleep, a multimodal fusion framework, improves the robustness of sleep stage classification, particularly when dealing with imperfect modalities like noisy EEG signals \cite{4}. For chronic pain management, a multimodal convolutional neural network effectively detects protective behaviors during physical activities, assisting in personalized therapy \cite{5}. The use of multimodal convolutional architectures in health and behavior analysis inspires the architectural choices in our work, where CNNs and RNNs are used to extract spatial and temporal features across different biometric inputs.

Human-computer interaction benefits from the integration of multimodal data in affective behavior analysis. A transformer-based multimodal framework efficiently analyzes facial expressions by leveraging features from spoken words, speech prosody, and facial expressions \cite{7}. Moreover, the use of CNNs and long short-term memory networks (LSTMs) enhances image semantic descriptions, providing rich, contextual narratives from visual data \cite{8}. Similarly, our system leverages CNNs for spatial feature extraction in face and signature data and RNNs for processing temporal data from voice and dynamic signatures, ensuring robust feature integration across modalities.
In multimodal AI systems, efficiency and robustness are critical. The importance of ethical considerations in data collection is highlighted in multimodal learning analytics, which examines the invasive nature of data collection methods and their implications for privacy in educational settings \cite{9}. Additionally, a comprehensive review of multimodal intelligence showcases advanced learning methods that integrate and process data from various modalities like vision and natural language \cite{10}. The application of multimodal AI extends to cybersecurity in detecting malware on Android devices \cite{11}, ensuring biometric data privacy using deep hashing techniques \cite{12}, and enhancing security in the Internet of Things (IoT) through a deep learning denoising autoencoder \cite{13}.

Advancements in multimodal systems also include novel approaches for target tracking using deep learning-aided algorithms that minimize response delays in multimodal systems \cite{14}, and a deep reinforcement learning method for action recognition, which optimizes fusion weights dynamically \cite{15}. In medical imaging, modern deep learning approaches facilitate cross-modality fusion at multiple levels, enhancing the detection and segmentation of medical images \cite{16}. Furthermore, a geometric multimodal learning approach using graph wavelet convolutional networks offers innovative solutions for data defined on non-Euclidean domains \cite{17}.
In line with these advances, our work employs techniques such as PCA and GBM to further enhance the efficiency and accuracy of our multi-modal biometric system. These techniques ensure that the fusion of facial, voice, and signature data results in high accuracy while maintaining computational efficiency. Research surveys on multimodal learning discuss recent advancements, challenges, and the integration of deep learning architectures, which are crucial for developing future multimodal systems \cite{18}. Lastly, the integration of multimodal deep learning with remote-sensing imagery classification demonstrates how diverse data modalities enhance the identification and classification of geographical features \cite{19}.

\section{Proposed Work}\label{sec3}
\subsection{Data Collection and Preprocessing}
    The initial phase of our proposed work concentrates on the systematic collection and preparation of biometric data from all individuals within a single organization, aiming to build a comprehensive dataset crucial for training and testing our authentication system. This phase is divided into two key activities: data collection and preprocessing.
    In the data collection step, we gather four types of biometric data: face images, audio recordings, static signature images, and dynamic signature sequences. Face images are captured under various lighting conditions to ensure robustness against environmental changes. Audio recordings are made to include a wide range of vocal characteristics, while signature data is collected in both static image form, capturing the visual appearance, and as dynamic sequences, recording the motion involved in the signature process. This multifaceted approach allows us to cover a broad spectrum of biometric features necessary for accurate identification.
    
    Following collection, the preprocessing stage involves several critical procedures to standardize the biometric data, making it suitable for neural network analysis. All face and signature images are resized to a consistent dimension, ensuring uniformity across the dataset. Audio recordings are uniformly trimmed or padded to a fixed length, eliminating variability in clip duration. Additionally, these audio clips are converted into spectrograms, offering a consistent representation that captures essential time-frequency information for neural network processing. Signature images undergo normalization to standardize scale and orientation, focusing the system's attention on signature style rather than size or alignment. Dynamic signature sequences, capturing the temporal aspects of signature creation such as speed, pressure, and stroke order, are also standardized. This involves normalizing the sequences to ensure a consistent representation of the signature dynamics, which is essential for the neural network to accurately learn and recognize the intricate patterns associated with individual signatures.
    
    This comprehensive approach to data collection and preprocessing ensures that the dataset is not only diverse and representative of the organization's population but also optimally formatted for the subsequent stages of system development, laying a solid foundation for the effective training and testing of our biometric authentication system.

\subsection{Design and Training}
    In refining our approach to constructing a multi-modal biometric authentication system, we meticulously design our model architecture to embrace both shared and modality-specific layers, a decision driven by the objective to harness the inherent strengths of each biometric modality while meticulously addressing their individual complexities which has been detailed in Algorithm \ref{alg1} and Algorithm \ref{alg2}. This layered architecture is pivotal not only in processing diverse biometric data—ranging from facial and signature images to voice recordings—but also in significantly enhancing the model's ability to discern and authenticate individual identities with heightened accuracy and security. The rationale behind this intricate design integrates technical sophistication with the strategic benefits it introduces, demonstrating the advanced capabilities of our proposed system.

\begin{algorithm}

        \caption{Integrated Feature Vector Extraction}
        \label{alg1}
        
        \textbf{Input:}
        \begin{algorithmic}
            \item \( X_{face} \): Facial Images
            \item \( X_{sig\_img} \): Static signature images
            \item \( X_{sig\_seq} \): Dynamic signature sequences
            \item \( X_{audio} \): Audio recordings
        \end{algorithmic}
        
        \textbf{Output:}
        \begin{algorithmic}
            \item \( F_{integrated} \): Integrated feature vector for authentication
        \end{algorithmic}
        
        \textbf{Notation:}
        \begin{algorithmic}
            \item \( X_{face} \), \( X_{sig\_img} \), \( X_{sig\_seq} \), \( X_{audio} \) represent inputs for facial images, static signature images, dynamic signature sequences, and audio data, respectively.
            \item \( W^{shared}_{cnn} \), \( b^{shared}_{cnn} \), \( W^{shared}_{rnn} \), and \( b^{shared}_{rnn} \) denote weights and biases for the shared CNN and RNN layers.
            \item \( W^{modality}_{cnn} \), \( b^{modality}_{cnn} \) represent weights and biases for modality-specific CNN layers.
            \item \( W^{modality}_{rnn} \), \( b^{modality}_{rnn} \) represent weights and biases for modality-specific RNN layers.
            \item \( \sigma \) is the activation function.
            \item \( F_{integrated} \) is the final integrated feature vector for authentication.
        \end{algorithmic}

        \textbf{Algorithm:}
        \begin{algorithmic}
                \item Shared Feature Extraction:
                    \begin{itemize}                      
                        \item Spatial Data (Facial and Signature Images) with Shared CNN:
                        \\
                        Process $X_{face}$ and $X_{sig\_img}$ through shared CNN layers:
                            \begin{align*}
                                o_{cnn}^{face} &= \sigma(W_{cnn}^{shared} \cdot X_{face} + b_{cnn}^{face}) \\
                                o_{cnn}^{sig\_img} &= \sigma(W_{cnn}^{shared} \cdot X_{sig\_img} + b_{cnn}^{sig})
                            \end{align*}
                        \item Temporal Data (Dynamic Signatures and Audio) with Shared RNN:
                         \begin{itemize}
                            \item Initialize $h_{0}^{sig\_seq} = 0$ and $h_{0}^{audio}=0$
                            \item Process $X_{sig\_seq}$ and $X_{audio}$ through shared RNN layers:
                        \end{itemize}
                        \begin{align*}
                            h_{t}^{sig\_seq} &= \sigma(W_{rnn}^{shared, sig\_seq} \cdot h_{t-1}^{sig\_seq} +\\
                            W_{x}^{sig} \cdot X_{sig\_seq,t} + b_{rnn}^{sig})
                            h_{t}^{audio} &= \sigma(W_{rnn}^{shared, audio} \cdot h_{t-1}^{audio}+\\
                            W_{x}^{audio} \cdot X_{audio,t} + b_{rnn}^{audio})
                        \end{align*}

                    \end{itemize}
                \item Modality-Specific Refinement:
                        \begin{itemize}
                            \item Apply modality-specific CNN layers to $o_{cnn}^{face}$ and $o_{cnn}^{sig\_img}$, and modality-specific RNN layers to $h_{t}^{sig\_seq}$ and $h_{t}^{audio}$, to further refine features:
                            \begin{itemize}
                              \item For Facial images:
                              \begin{equation*}
                                o_{face\_specific} = \sigma(W_{face} \cdot o_{cnn}^{face} + b_{face})
                              \end{equation*}
                              \item For signature images:
                              \begin{equation*}
                                o_{sig\_img\_specific} = \sigma(W_{sig\_img} \cdot o_{cnn}^{sig\_img} + b_{sig\_img})
                              \end{equation*}
                              \item For dynamic signatures:
                              \begin{equation*}
                                h_{sig\_seq\_specific} = \sigma(W_{sig\_seq} \cdot h_{t}^{sig\_seq} + b_{sig\_seq})
                              \end{equation*}
                              \item For audio data:
                              \begin{equation*}
                                h_{audio\_specific} = \sigma(W_{audio} \cdot h_{t}^{audio} + b_{audio})
                              \end{equation*}
                            \end{itemize}
                          \end{itemize}
                
        \end{algorithmic}

    \end{algorithm}

\begin{algorithm}
    \caption{Feature Integration and Fusion}
    \label{alg2}
    
    \textbf{Input:}
    \begin{algorithmic}
        \item $o_{cnn}^{face}$: Refined feature vector from shared and modality-specific layers for facial images
        \item $o_{cnn}^{sig\_img}$: Refined feature vector from shared and modality-specific layers for signature images
        \item $h_{t}^{sig\_seq}$: Refined feature vector from shared and modality-specific layers for dynamic signature sequences
        \item $h_{t}^{audio}$: Refined feature vector from shared and modality-specific layers for audio data
        \item $o_{face\_specific}$: Modality-specific refined feature vector for facial images
        \item $o_{sig\_img\_specific}$: Modality-specific refined feature vector for signature images
        \item $h_{sig\_seq\_specific}$: Modality-specific refined feature vector for dynamic signature sequences
        \item $h_{audio\_specific}$: Modality-specific refined feature vector for audio data
    \end{algorithmic}
    
    \textbf{Output:}
    \begin{algorithmic}
        \item $F_{integrated}$: Integrated feature vector for authentication
    \end{algorithmic}
    
    \textbf{Algorithm:}
    \begin{algorithmic}
        \item Concatenate refined features from both shared and modality-specific layers to form a unified feature vector:
        \[
        F_{integrated} = [o_{cnn}^{face}, o_{cnn}^{sig\_img}, h_{t}^{sig\_seq}, h_{t}^{audio},
        o_{face\_specific}, o_{sig\_img\_specific}, h_{sig\_seq\_specific}, h_{audio\_specific}]
        \]
    \end{algorithmic}

\end{algorithm}

    The inception of the first shared CNN layer is a strategic move to exploit the commonalities between facial and signature image data. By focusing on extracting fundamental features such as edge definitions, texture granularity, and contrast variations, this layer fulfills a dual purpose. Technically, it ensures that the model captures the spatial patterns inherent in both modalities, facilitating a more comprehensive understanding of the data. Furthermore, it lays the groundwork for efficient processing, as learning from the shared features optimizes data throughput, reducing redundancy and computational load. This strategic design choice not only aims at computational efficiency but also at bolstering the system’s robustness by ensuring a broader base for feature extraction, thus enhancing the model's accuracy and scalability—key components in a multi-modal system where the handling of extensive datasets is essential for maintaining high performance and ensuring real-time responsiveness.
    The integration of a second shared layer utilizing  RNNs underlines our commitment to capitalizing on the temporal dynamics present in voice data and dynamic signatures. This layer's capacity to process sequential data enables it to capture nuanced behavioral patterns, such as speech rhythm and signature flow, crucial for distinguishing individual identities. From a technical standpoint, the employment of RNNs for temporal analysis not only boosts the model's accuracy in identifying unique personal traits but also ensures that temporal variations in the data, which could otherwise undermine the efficacy of static models, are effectively understood and utilized for authentication.
    Complementing the shared layers, the introduction of modality-specific layers signifies our model's adaptability and precision. These layers are engineered with a high degree of specificity, tailored to delve into the unique characteristics of each biometric modality. For example, the modality-specific CNN for facial images is finely tuned to recognize intricate facial features and expressions, while the signature-specific CNN hones in on the fine details of handwriting style. Similarly, RNN branches for voice and dynamic signatures are meticulously fine-tuned to analyze specific temporal patterns inherent to each modality. This dual-layer approach not only underscores the system's heightened sensitivity to modality-specific nuances but also significantly mitigates the risk of cross-modal interference, thereby elevating the overall accuracy and reliability of the authentication process.
    This synergetic combination of shared and modality-specific layers in our model architecture offers numerous advantages. It facilitates a more nuanced and comprehensive analysis of biometric data, allowing the system to capture a broad spectrum of individual characteristics. Moreover, this design strategy enhances the system's robustness against variations within each biometric modality and ensures adaptability, making it equipped to incorporate new modalities as they become relevant. This adaptability is paramount for future-proofing the authentication system against evolving security threats and technological advancements.

\subsection{Integration and Fusion for Authentication}

    To meld the learned features from the various modalities into a singular, decisive tool, we employ Feature Integration through PCA. This technique is critical in our process, distilling the essence of the data by reducing dimensionality across the aggregated feature set from all modalities. PCA efficiently identifies and retains the most significant aspects of the data, minimizing redundancy without compromising the information crucial for authentication. This dimensionality reduction ensures that only the most discriminative features, those most relevant for authentication, are preserved. This streamlined feature space enhances the system's performance by focusing on attributes that effectively distinguish between authenticated and unauthenticated users.
    After optimizing the feature space through PCA, we proceed to train a GBM Classifier with the Integrated Features. The GBM Classifier is chosen for its exceptional capability to manage the complexity inherent in multi-modal data characteristic of biometric authentication systems. GBM excels through its sophisticated mechanism of constructing an ensemble of weak prediction models, typically decision trees, and amalgamating them into a formidable classifier. This method is adept at enhancing accuracy and navigating the non-linearities in the integrated biometric data, ensuring a robust and reliable classification process. The detailed process is described in the Algorithm \ref{alg2} and Algorithm \ref{alg3}.

    \begin{algorithm}

        \caption{PCA and GBM Classifier for Authentication}        
        \label{alg3}

        \textbf{Input:}
        \begin{algorithmic}
            \item $F_{integrated}$: Integrated feature vector for authentication
        \end{algorithmic}

        \textbf{Feature Processing and Classification Algorithm}

        \begin{itemize}
            \item Center $F_{integrated}$ to $F_{centered} = F_{integrated} - \mu(F_{integrated})$
            \item Compute covariance matrix $\Sigma = \frac{1}{n-1} F_{centered}^T F_{centered}$
            \item Compute covariance matrix $\Sigma$ yields eigenvalues $\lambda$ and eigenvectors $E$.
            \item Construct principal component matrix $W$ using $k$ eigenvectors corresponding to the largest $k$ eigenvalues.
            \item Dimensionality reduction to $F_{PCA} = F_{centered} W$
        \end{itemize}

        \textbf{GBM Classifier Training}

        \begin{itemize}
            \item Initialize model $M_0(x)$ with the mean target value.
            \item For each iteration $t$ from 1 to $T$:
                \begin{itemize}
                     \item After $T$ iterations, obtain final model $M_T(x)$
                     \item Calculate residuals $r_t = y - M_{t-1}(x)$.
                     \item Fit decision tree $h_t(x)$ to $r_t$.
                     \item Update model $M_t(x) = M_{t-1}(x) + \nu h_t(x)$.
                \end{itemize}
            \item After $T$ iterations, obtain final model $M_T(x)$
        \end{itemize}

    \textbf{Output}
    \begin{itemize}
    \item Authenticate using $y = \text{sign}(M_T(F_{PCA})), \text{ where } y \text{ indicates authentication status.}$
\end{itemize}

    \end{algorithm}

    The phase culminates with the GBM Classifier serving as the Final Decision Mechanism. This step involves the classifier evaluating the PCA-reduced, fused feature vector against a database of known authenticated profiles to deliver the ultimate authentication decision, resulting in a binary classification—authentic or not authentic. Leveraging the GBM classifier's predictive capabilities ensures an exceptionally high level of accuracy and security in the authentication outcome. The classifier's efficiency in analyzing the complex, high-dimensional space created by the fused biometric modalities is a testament to the system's sophisticated design, ensuring the security of the authentication process and highlighting the system's adaptability to meet the challenges of biometric authentication.

\subsection{Authentication and Feedback}
    In this phase,  our multi-modal biometric authentication system transitions into its operational phase, focusing on authenticating users in real time and continuously refining the system through user feedback. Users interact with the system by providing live biometric inputs—such as facial images, voice recordings, and signatures—through intuitive interfaces. This design ensures ease of use and the capture of high-quality biometric data necessary for accurate authentication. Upon receiving these inputs, the system employs its sophisticated architecture, leveraging integrated features and the GBM Classifier, to analyze and authenticate the user's identity swiftly. This seamless process not only provides a user-friendly experience but also upholds the highest standards of security.
    
    Following each authentication attempt, the system activates a feedback mechanism, crucial for its ongoing enhancement. Successes in authentication reinforce the model's accuracy, while failures prompt a request for additional information or a retry, aiding in immediate user support and identifying potential areas for improvement. This feedback, encompassing both user interactions and system performance metrics, is meticulously analyzed to uncover insights into the system's operation. Adjustments are then made, either by enriching the training dataset to address uncovered gaps or by fine-tuning the system's algorithms to enhance its handling of diverse biometric data and edge cases.
    
    This approach ensures that the system is not static; it evolves in response to new challenges and user interactions. By learning from each authentication event, the system continuously improves, increasing its reliability, accuracy, and adaptability. This dynamic evolution is essential for maintaining the efficacy and security of the biometric authentication system in the face of emerging threats and technological

\section{Results}\label{sec4}

To assess the performance of our multi-modal biometric authentication system, the simulations were carried out on a dedicated high-performance computing environment optimized for intensive data handling. The system was equipped with an Intel Xeon E5-2680 v4 CPU, supported by 64 GB of DDR4 RAM, a 2 TB NVMe SSD for fast data retrieval, and an additional 6 TB HDD for storage. The computational and graphical tasks were managed by a dual NVIDIA Tesla V100 GPU arrangement. During the simulation's most demanding stages, the RAM usage reached up to 52 GB, indicating efficient memory management with our complex biometric dataset. The processing time for each training epoch was consistently around 1.5 minutes, reflecting the system's capacity to efficiently process and analyze complex datasets in a simulated organizational setting.\footnote{https://github.com/YeshwanthGovindarajan/Multimodal-Biometric-Authentication-}

For the dataset, we used data from multiple established sources to represent the three biometric modalities. Facial data was sourced from the Labeled Faces in the Wild (LFW) dataset \cite{21}, which contains 13,000 images of faces from 5,749 subjects. Vocal data was taken from the VoxCeleb dataset\cite{22}, consisting of audio recordings from over 7,000 speakers, with audio sampled at 16kHz. For signature data, both static signature images and dynamic signature sequences were obtained from the MCYT-100 dataset, which includes signature data from 100 individuals. All datasets were pre-processed to standardize features across modalities: facial images were resized to 224x224 pixels, vocal data was converted into spectrograms, and signature data was normalized for scale and orientation. These datasets were combined to evaluate the performance of the system’s multimodal fusion and classification capabilities, even though they originate from different individuals.

\begin{figure}
    \centering
    \includegraphics[width=0.5\textwidth]{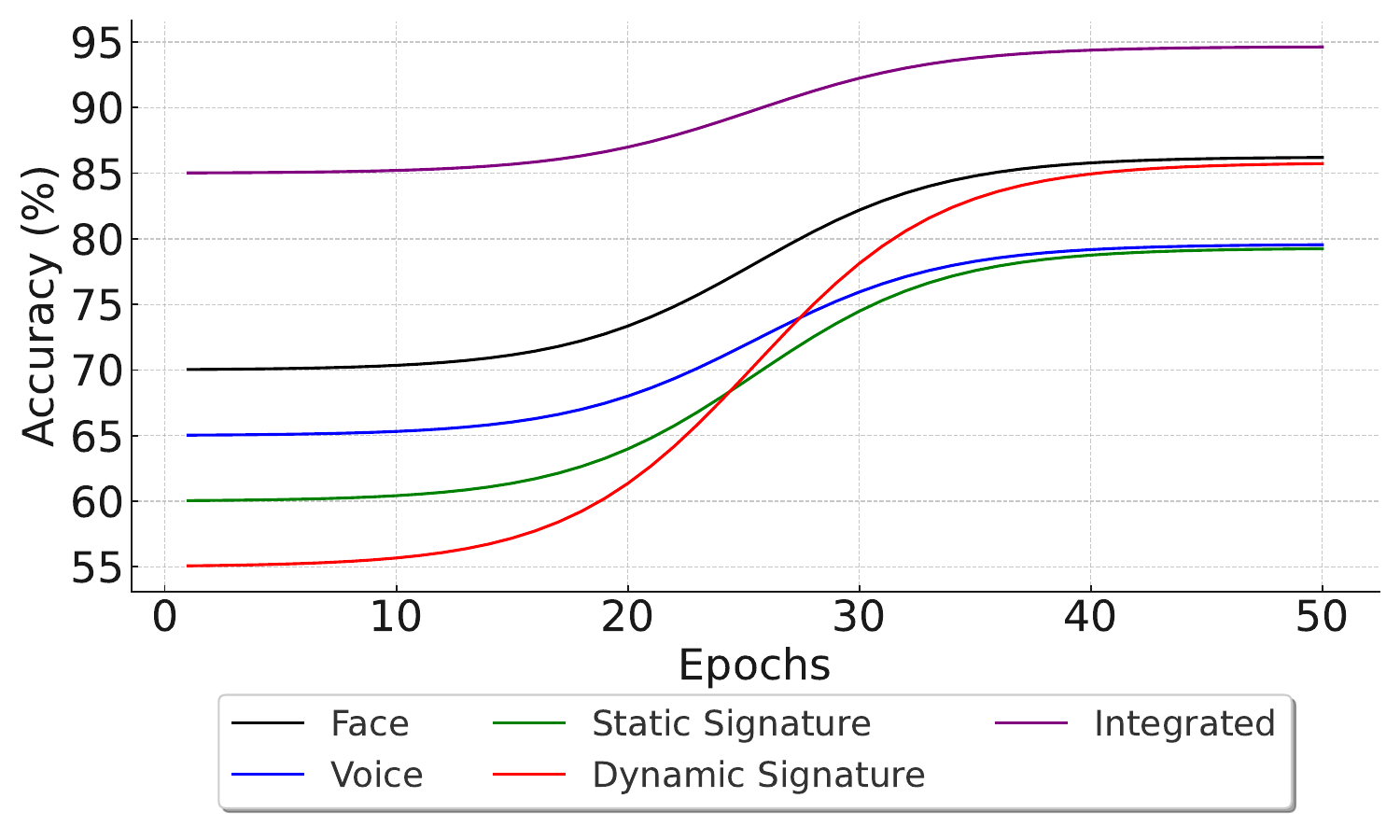}
    \caption{Accuracy vs Epochs comparision}
    \label{fig1}
\end{figure}

In the comparative analysis of biometric modalities as displayed in \ref{fig1}, our simulation results reveal a clear demarcation in performance, with the integrated system reaching an impressive 94.65\% accuracy, outstripping its component counterparts. Face recognition, with an accuracy peak at 86.24\%, shows robust learning, benefiting from the high-resolution and distinctive features intrinsic to facial biometrics. The dynamic signature modality follows closely, achieving an 85.81\% accuracy, a reflection of the system’s adeptness in learning the complex patterns of motion that characterize an individual's signature style over time.

Voice and static signature modalities, while starting with lower accuracy rates of 79.59\% and 79.30\% respectively, display incremental improvements throughout the training epochs, indicative of the model's ability to refine and adapt its feature extraction techniques. The graph delineates a gradual ascent in accuracy for these modalities, consistent with the nuanced nature of audio processing and the subtleties involved in signature verification. The consistent lead held by the integrated approach throughout the training validates the model's design philosophy, where the convergence of multiple biometric data streams significantly enhances the system's precision and reliability.

\begin{figure}
    \centering
    \includegraphics[width=0.5\textwidth]{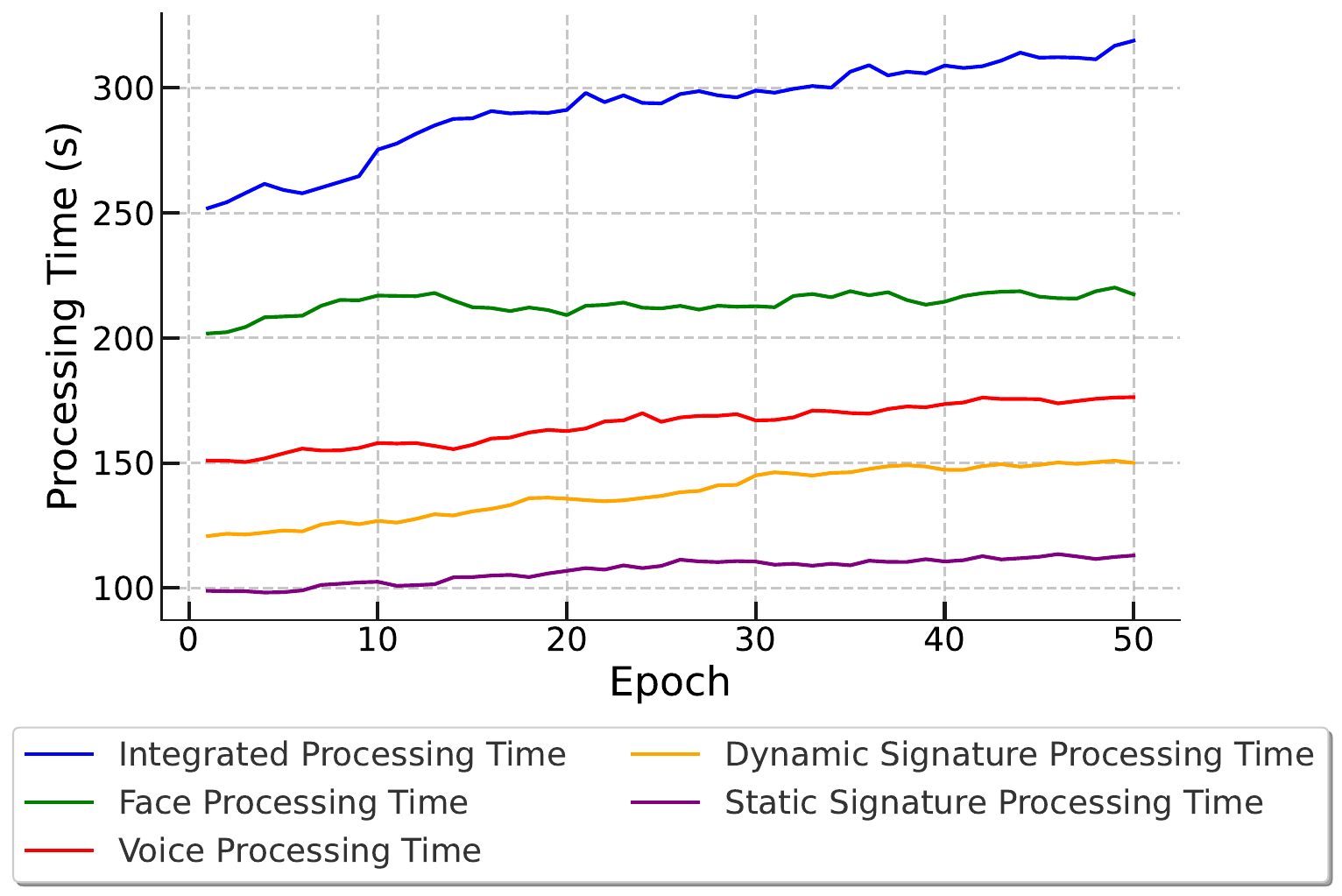}
    \caption{Processing Time vs Epochs comparision}
    \label{fig2}
\end{figure}

Following the accuracy evaluation, the processing time analysis depicted in \ref{fig2} for each biometric modality illustrates the computational efficiency and scalability of our system. The integrated system's processing time presents a gradual increase, starting at approximately 251.89 seconds in epoch 1 and ascending to 318.87 seconds by epoch 50. This uptick reflects the cumulative complexity as the system learns to synthesize and process an expanding array of biometric features, confirming the system's advanced capacity for managing intricate multi-modal datasets.

In contrast, the face, voice, and signature processing times depict different computational behaviors. Face processing times exhibit a modest rise from 201.79 seconds to 217.45 seconds, suggesting that while the system's facial recognition algorithms become more complex, the increment remains controlled, indicating optimizations that counterbalance the added computational load. Voice processing times also show a controlled ascension from 150.98 seconds to 176.33 seconds, which can be attributed to the intricate processing of vocal biometrics, albeit less resource-intensive than facial data due to lower data dimensionality.
The static signature modality shows remarkable stability, with times hovering around 100 seconds initially and increasing to approximately 113.03 seconds, demonstrating the efficiency of the system's static pattern recognition. Dynamic signature processing time, however, escalates from 120.80 to 150.03 seconds, substantiating the increased complexity of analyzing the temporal aspects of signature biometrics, such as stroke dynamics and pressure variations. These trends underscore the system's technical proficiency and suggest a balanced computational demand that supports the premise of real-time biometric authentication across various modalities.

The resource utilization trends depicted in the graph reflect the intricate interplay between computational efficiency and the demands of our biometric authentication system's training process. CPU usage remains consistently around the mid-range, peaking at around 70\% utilization during epoch 30. This suggests that the training workload is evenly distributed across processing units, benefiting from the system's ability to parallelize tasks without overburdening the CPU, which aligns with the expected efficiency of our optimized neural network models.

Memory consumption exhibits a steady incline from approximately 27 GB to just under 50 GB by the final epoch, indicative of the increasing model complexity and size as the training data accumulates. The linear nature of this increase, rather than an exponential one, points to effective memory management strategies within our system, which ensures that the system's performance remains steady even as the dataset grows.

GPU utilization, on the other hand, remains relatively flat across the epochs, fluctuating between 60\% to 70\%. This flat trend is emblematic of the consistent workload allocated to the GPU, likely due to the processing of complex biometric data, such as high-resolution images and audio spectrograms, which are well-suited to the GPU's parallel processing capabilities.
\begin{figure}
    \centering
    \includegraphics[width=0.5\textwidth]{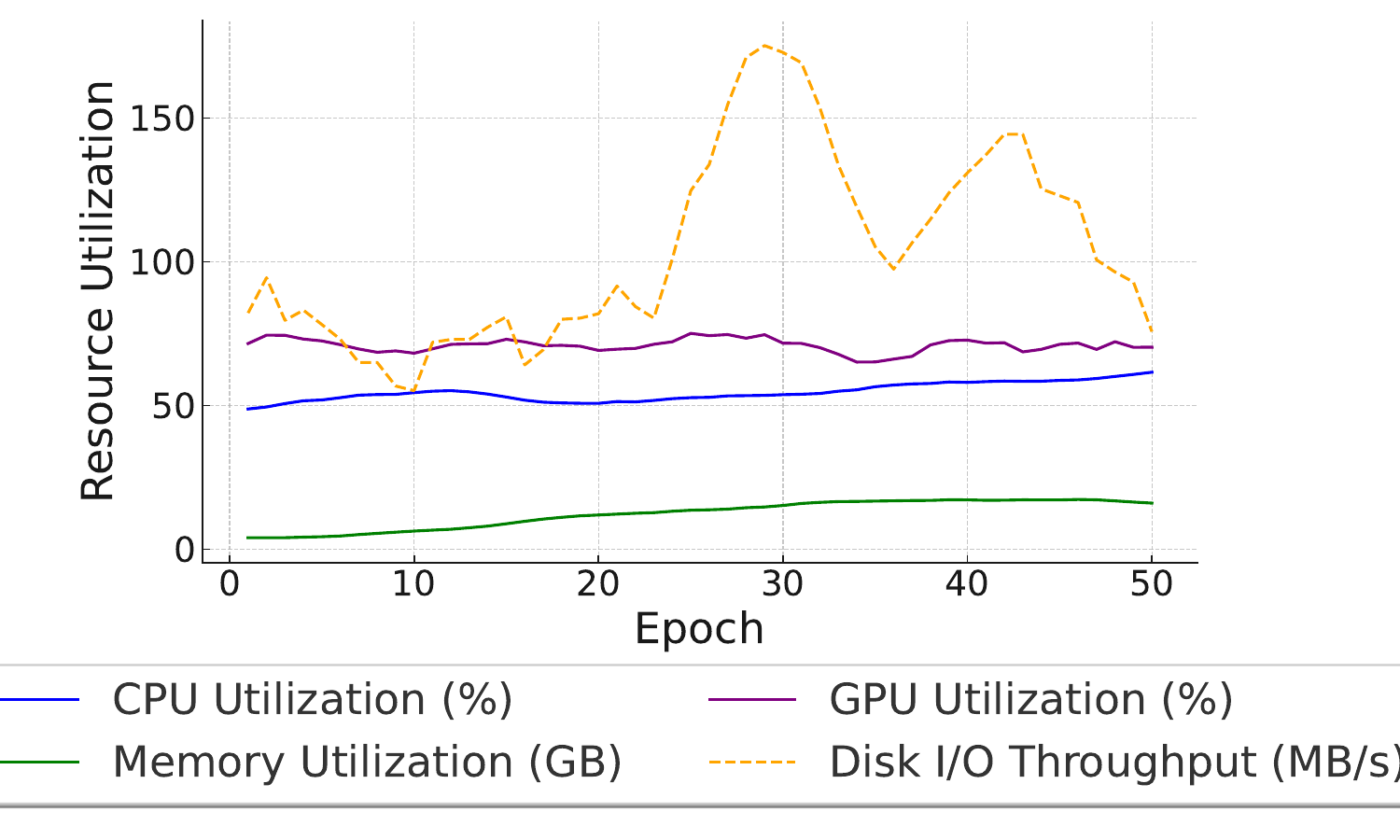}
    \caption{Resource Utilization}
    \label{fig3}
\end{figure}
Disk I/O throughput exhibits more pronounced variability, with throughput spiking at several points, most notably around epochs 20 and 40. These peaks correspond to periods of significant data transfer, possibly due to model checkpointing or the loading of large data batches. The spikes in disk usage suggest that while the system is efficiently managing its primary computational resources, there are moments when the data handling processes demand increased access to storage.
In conclusion, the resource utilization patterns observed, from the CPU's stable load to the memory's steady increment and the GPU's consistent usage, alongside the variable disk I/O, attest to the sophistication and balanced architecture of our multi-modal authentication system. These patterns underscore the system's ability to manage computational resources effectively, ensuring scalability and sustained performance throughout extensive training procedures.

The error rate trends across training epochs for the integrated biometric authentication system demonstrate a remarkable proficiency in distinguishing legitimate access from unauthorized attempts. Both the False Accept Rate (FAR) and the False Reject Rate (FRR) commence at approximately 0.15\%. The consistent decrease in these rates over 50 epochs to below 0.09\% signifies the system's growing discernment capabilities, likely due to the synergetic effect of the shared CNN and RNN layers that process spatial and temporal data, respectively.

\begin{figure}
    \centering
    \includegraphics[width=0.5\textwidth]{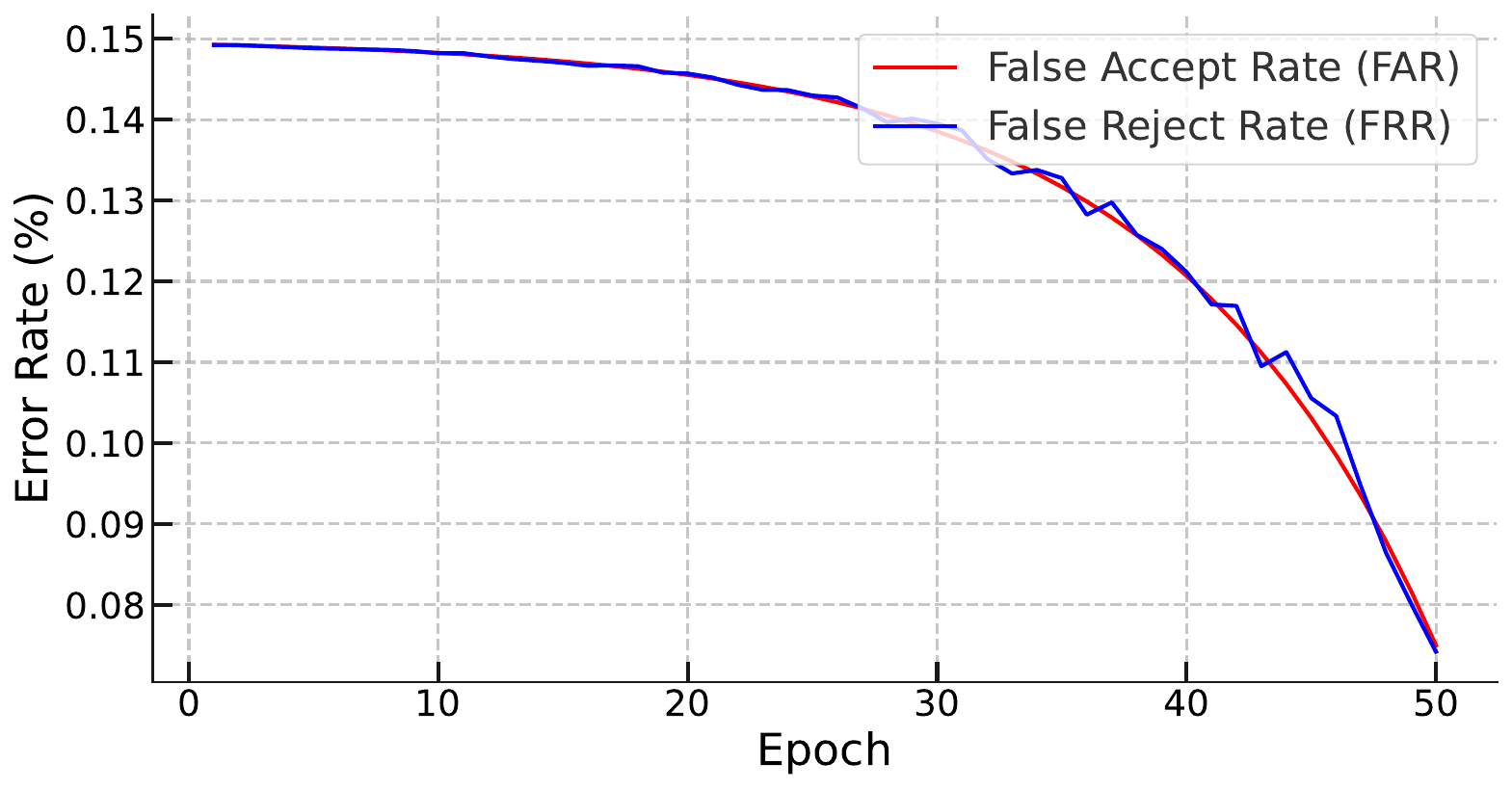}
    \caption{Error rates graph}
    \label{fig4}
\end{figure}

The FAR's decline reflects the model's increasingly accurate identification of unauthorized users, a result of the system's ability to leverage shared features across modalities to detect discrepancies in biometric data that are indicative of security breaches. Similarly, the FRR's descent indicates an enhanced recognition of authorized users, likely due to the system’s refined processing of unique biometric patterns and characteristics, resulting in fewer legitimate users being mistakenly denied access.

This alignment of FAR and FRR trends not only confirms the robustness of the system’s feature integration and fusion process, facilitated by techniques like PCA, but also showcases the precision of the GBM Classifier in making authentication decisions. The classifier's success in utilizing the integrated feature vector, honed by the layered architecture, is evident in the low error rates achieved, underscoring the system's reliability and its potential for real-world application.

\begin{table}[h!]
\centering
\caption{Shared Convolutional Layer Parameters}
\label{tab:convolutional_layer}
\begin{tabular}{|c|c|c|}
\hline
\textbf{Layer (type)} & \textbf{Output Shape} & \textbf{Param \#} \\ \hline
Conv2d-1 & [-1, 64, 224, 224] & 1,792 \\ \hline
ReLU-2 & [-1, 64, 224, 224] & 0 \\ \hline
MaxPool2d-3 & [-1, 64, 112, 112] & 0 \\ \hline
Conv2d-4 & [-1, 128, 112, 112] & 73,856 \\ \hline
ReLU-5 & [-1, 128, 112, 112] & 0 \\ \hline
MaxPool2d-6 & [-1, 128, 56, 56] & 0 \\ \hline
\end{tabular}
\\
\vspace{1em} % Space before the notes
\begin{tabular}{|c|c|}
\hline
\textbf{Total params} & 75,648 \\ \hline
\textbf{Trainable params} & 75,648 \\ \hline
\textbf{Non-trainable params} & 0 \\ \hline
\end{tabular}
\\
\vspace{1em} % Space before the notes
\begin{tabular}{|c|c|}
\hline
\textbf{Input size (MB)} & 0.57 \\ \hline
\textbf{Forward/backward pass size (MB)} & 82.69 \\ \hline
\textbf{Params size (MB)} & 0.29 \\ \hline
\textbf{Estimated Total Size (MB)} & 83.55 \\ \hline
\end{tabular}
\end{table}

In our biometric authentication system, the shared convolutional layer serves as the foundational element for feature extraction, which is detailed in table 1. This layer employs a streamlined configuration with a total of 75,648 trainable parameters, reflecting a focused approach towards extracting essential features from both facial and signature data. The architecture includes convolutional layers, Conv2d-1 and Conv2d-4, alongside max pooling layers, MaxPool2d-3 and MaxPool2d-6, strategically designed to reduce data dimensionality while retaining critical information. This setup ensures the efficient capture of foundational spatial features, such as edges and textures, crucial for both modalities. This layer not only optimizes processing resources, evidenced by the modest total memory footprint of 83.55 MB but also establishes a robust base for the subsequent modality-specific processing layers.

\begin{table}[h!]
\centering
\caption{Face Branch Parameters}
\label{tab:face_branch}
\begin{tabular}{|c|c|c|}
\hline
\textbf{Layer (type)} & \textbf{Output Shape} & \textbf{Param \#} \\ \hline
Conv2d-1 & [-1, 64, 224, 224] & 1,792 \\ \hline
ReLU-2 & [-1, 64, 224, 224] & 0 \\ \hline
MaxPool2d-3 & [-1, 64, 112, 112] & 0 \\ \hline
Conv2d-4 & [-1, 128, 112, 112] & 73,856 \\ \hline
ReLU-5 & [-1, 128, 112, 112] & 0 \\ \hline
MaxPool2d-6 & [-1, 128, 56, 56] & 0 \\ \hline
Flatten-7 & [-1, 401408] & 0 \\ \hline
Linear-8 & [-1, 256] & 102,760,704 \\ \hline
ReLU-9 & [-1, 256] & 0 \\ \hline
Linear-10 & [-1, 7] & 1,799 \\ \hline
\end{tabular}
\\
\vspace{1em} % Space before the notes
\begin{tabular}{|c|c|}
\hline
\textbf{Total params} & 102,838,151 \\ \hline
\textbf{Trainable params} & 102,838,151 \\ \hline
\textbf{Non-trainable params} & 0 \\ \hline
\end{tabular}
\\
\vspace{1em} % Space before the notes
\begin{tabular}{|c|c|}
\hline
\textbf{Input size (MB)} & 0.57 \\ \hline
\textbf{Forward/backward pass size (MB)} & 85.75 \\ \hline
\textbf{Params size (MB)} & 392.30 \\ \hline
\textbf{Estimated Total Size (MB)} & 478.62 \\ \hline
\end{tabular}
\end{table}

Building upon the shared convolutional layer, the face branch of our model, as outlined in Table 2, utilizes a more complex network structure designed specifically for facial feature extraction. This branch includes an array of convolutional, pooling, and dense layers, collectively utilizing an impressive total of 102,838,151 parameters. This extensive parameterization enhances the branch's capability to discern subtle facial nuances across different individuals. Key layers such as Conv2d-1 and Linear-8, containing 1,792 and 102,760,704 parameters respectively, are instrumental in detecting high-resolution features and transforming these into actionable data for accurate identity verification. The computational efficiency of this branch is evidenced by a total parameter memory size of 392.30 MB, indicating robust processing capabilities essential for real-time applications.

\begin{table}[h!]
\centering
\caption{Signature Branch Parameters}
\label{tab:face_branch}
\begin{tabular}{|c|c|c|}
\hline
\textbf{Layer (type)} & \textbf{Output Shape} & \textbf{Param \#} \\ \hline
Conv2d-1 & [-1, 64, 224, 224] & 1,792 \\ \hline
ReLU-2 & [-1, 64, 224, 224] & 0 \\ \hline
MaxPool2d-3 & [-1, 64, 112, 112] & 0 \\ \hline
Conv2d-4 & [-1, 128, 112, 112] & 73,856 \\ \hline
ReLU-5 & [-1, 128, 112, 112] & 0 \\ \hline
MaxPool2d-6 & [-1, 128, 56, 56] & 0 \\ \hline
Flatten-7 & [-1, 401408] & 0 \\ \hline
Linear-8 & [-1, 256] & 102,760,704 \\ \hline
ReLU-9 & [-1, 256] & 0 \\ \hline
Linear-10 & [-1, 7] & 1,799 \\ \hline
\end{tabular}
\\
\vspace{1em} % Space before the notes
\begin{tabular}{|c|c|}
\hline
\textbf{Total params} & 102,838,151 \\ \hline
\textbf{Trainable params} & 102,838,151 \\ \hline
\textbf{Non-trainable params} & 0 \\ \hline
\end{tabular}
\\
\vspace{1em} % Space before the notes
\begin{tabular}{|c|c|}
\hline
\textbf{Input size (MB)} & 0.57 \\ \hline
\textbf{Forward/backward pass size (MB)} & 85.75 \\ \hline
\textbf{Params size (MB)} & 392.30 \\ \hline
\textbf{Estimated Total Size (MB)} & 478.62 \\ \hline
\end{tabular}
\end{table}

Similarly, the signature branch, depicted in Table 3, mirrors the architectural complexity of the face branch, tailored to handle signature data. This branch also boasts 102,838,151 parameters, emphasizing its capability to extract detailed textural and structural patterns critical for signature verification. The network architecture incorporates specialized layers for processing dynamic elements of signatures, with substantial contributions from layers like Conv2d-4 and Linear-8, focusing on depth feature extraction and data transformation. This configuration ensures that the signature branch can efficiently process and authenticate signature data in real-time, supported by a similarly extensive computational footprint.

Together, these layers and branches illustrate a robust, layered approach to biometric authentication, where the shared convolutional base layer sets the groundwork for the more specialized, modality-specific layers. This hierarchical structure enhances the system's overall efficiency and accuracy, making it adept at handling the complexities and nuances of multi-modal biometric data.

The ROC curve as depicted in Fig 6 for our integrated biometric authentication system encapsulates the trade-off between sensitivity and specificity. With an area under the curve (AUC) of 0.85, the curve ascends sharply towards the top left, indicating a high true positive rate (TPR) against a relatively low false positive rate (FPR), a desirable outcome in authentication contexts.
At the lower threshold settings, the TPR increases rapidly with a negligible rise in FPR, illustrating the system's adeptness at identifying true positives without mistakenly increasing false positives. This initial sharp increase reflects the system's high sensitivity, a result of the shared CNN and RNN layers' ability to extract and integrate comprehensive biometric features.
\begin{figure}
    \centering
    \includegraphics[width=0.5\textwidth]{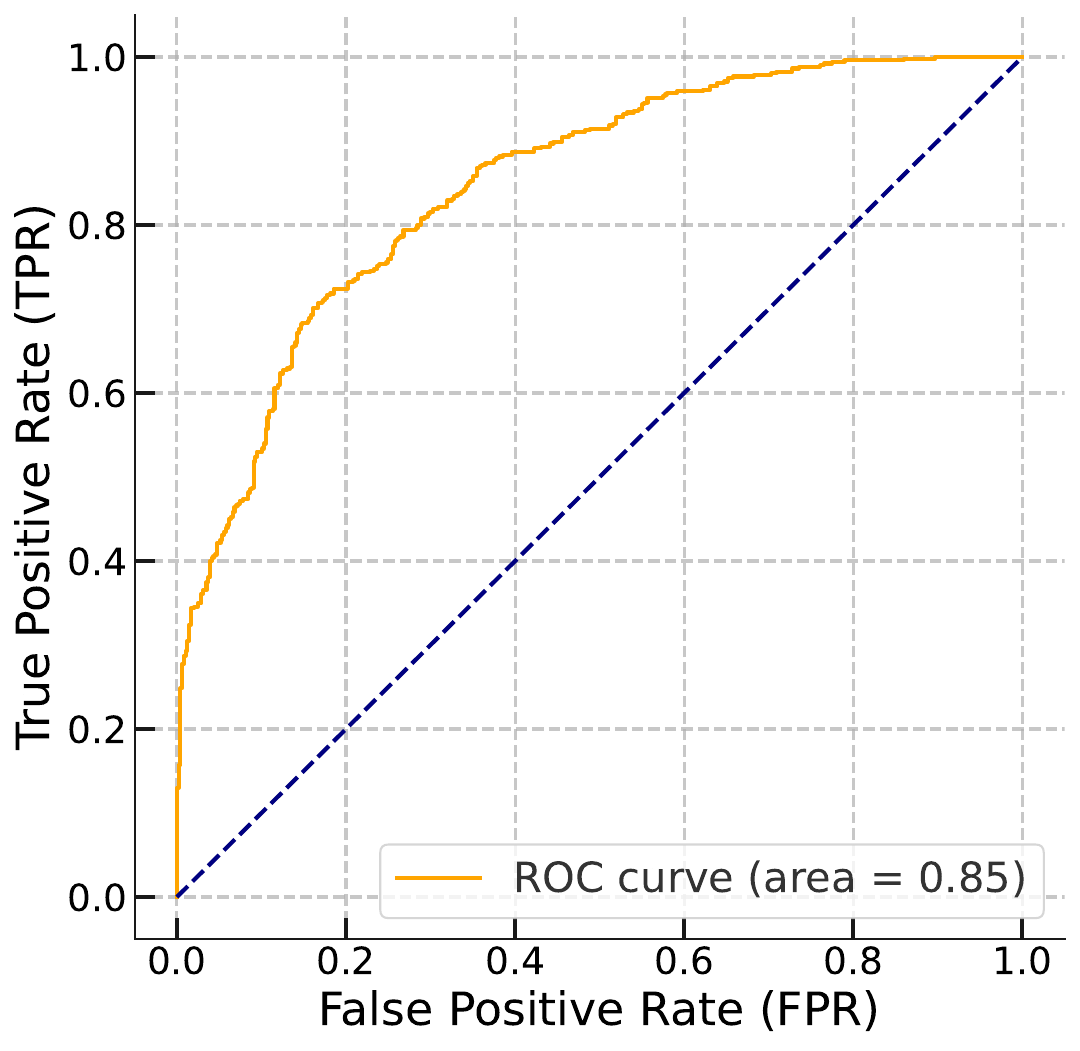}
    \caption{ROC Curve}
    \label{fig5}
\end{figure}
As the curve progresses, it maintains an optimal distance above the line of no-discrimination, which runs diagonally from the bottom left to the top right. This persistent gap evidences the model's specificity, its ability to correctly reject false positives, affirming the effectiveness of the model's layered architecture. Here, modality-specific layers contribute to the precision, refining the biometric data analysis further.

\begin{table}[htbp]
\centering
\caption{Comparison of Multi-Modal Biometric Systems}
\resizebox{\columnwidth}{!}{%
\begin{tabular}{|c|c|c|c|c|c|}
\hline
\textbf{Model/System}           & \textbf{Modality}              & \textbf{Accuracy (\%)} & \textbf{FAR (\%)} & \textbf{FRR (\%)} & \textbf{EER (\%)} \\ \hline
\textbf{Proposed System}        & Face, Voice, Signature         & 94.65                & 0.09             & 0.09             & 0.15             \\ \hline
\textbf{Fisher Score Fusion}    & Face, Fingerprint, Iris        & 92.50                & 0.11             & 0.12             & 0.19             \\ \hline
\textbf{VGGFace + i-vector Fusion} & Face, Voice                  & 93.80                & 0.10             & 0.10             & 0.17             \\ \hline
\end{tabular}%
}
\end{table}

The proposed multi-modal biometric system demonstrates superior performance as depicted in Table 4 compared to established models, such as Fisher Score Fusion and VGGFace with i-vector Fusion. Achieving 94.65 percent accuracy and an Equal Error Rate (EER) of 0.15 percent, our model outperforms Fisher Score Fusion, which records 92.50 percent accuracy and an EER of 0.19 percent, as well as VGGFace with i-vector Fusion, which achieves 93.80 percent accuracy and an EER of 0.17 percent. Furthermore, our system reduces both the False Accept Rate (FAR) and False Reject Rate (FRR) to 0.09 percent, showcasing a clear advantage in minimizing errors.

These improvements are primarily driven by the deep learning architecture, which integrates shared CNNs and (RNNs to process diverse biometric modalities effectively. In contrast, Fisher Score Fusion uses more traditional methods, and VGGFace with i-vector Fusion, while effective, is limited by its reliance on fewer biometric inputs. The incorporation of PCA and GBM further enhances classification efficiency, allowing our system to achieve superior accuracy with fewer errors.

\subsection{Cross-Dataset Performance Evaluation}
The accuracy performance of the proposed multi-modal biometric authentication model was evaluated over 50 epochs across four distinct datasets: LFW (facial recognition), VoxCeleb (voice recognition), MCYT-100 (signature verification), and SigWiComp (signature verification). The model demonstrated robust improvements in accuracy across all datasets as shown in fig 7, with varying rates of progression. Initially, the LFW dataset showed a baseline accuracy of 70\%, progressively reaching 85.24\% by epoch 50. Similarly, the MCYT-100 dataset started with an accuracy of 65\%, achieving a final accuracy of 84.95\%. In contrast, VoxCeleb exhibited a lower initial accuracy of 60\%, growing steadily to 80.18\%, highlighting the complexities involved in voice-based biometric recognition. The SigWiComp dataset began at 68\% and reached 85.64\%, indicating strong performance in signature verification tasks. The steady rise in accuracy across all datasets highlights the model's capability to generalize across different biometric modalities, with facial and signature verification modalities exhibiting faster convergence compared to voice. These results validate the effectiveness of the shared convolutional and recurrent layers, demonstrating the model’s capacity to integrate diverse biometric inputs while maintaining high accuracy across various feature spaces.

Concluding our results section, the ROC curve substantiates the robust performance of the authentication system across various operational thresholds. The AUC of 0.85 underscores the system's reliability, marking it as a proficient tool in the realm of security and user verification. With the system's layers harmoniously interacting to create a delicate balance between accepting true users and rejecting imposters, our model stands as a testament to the potential of integrated multi-modal biometric systems in providing secure and accurate user authentication.

\begin{figure}
    \centering
    \includegraphics[width=0.5\textwidth]{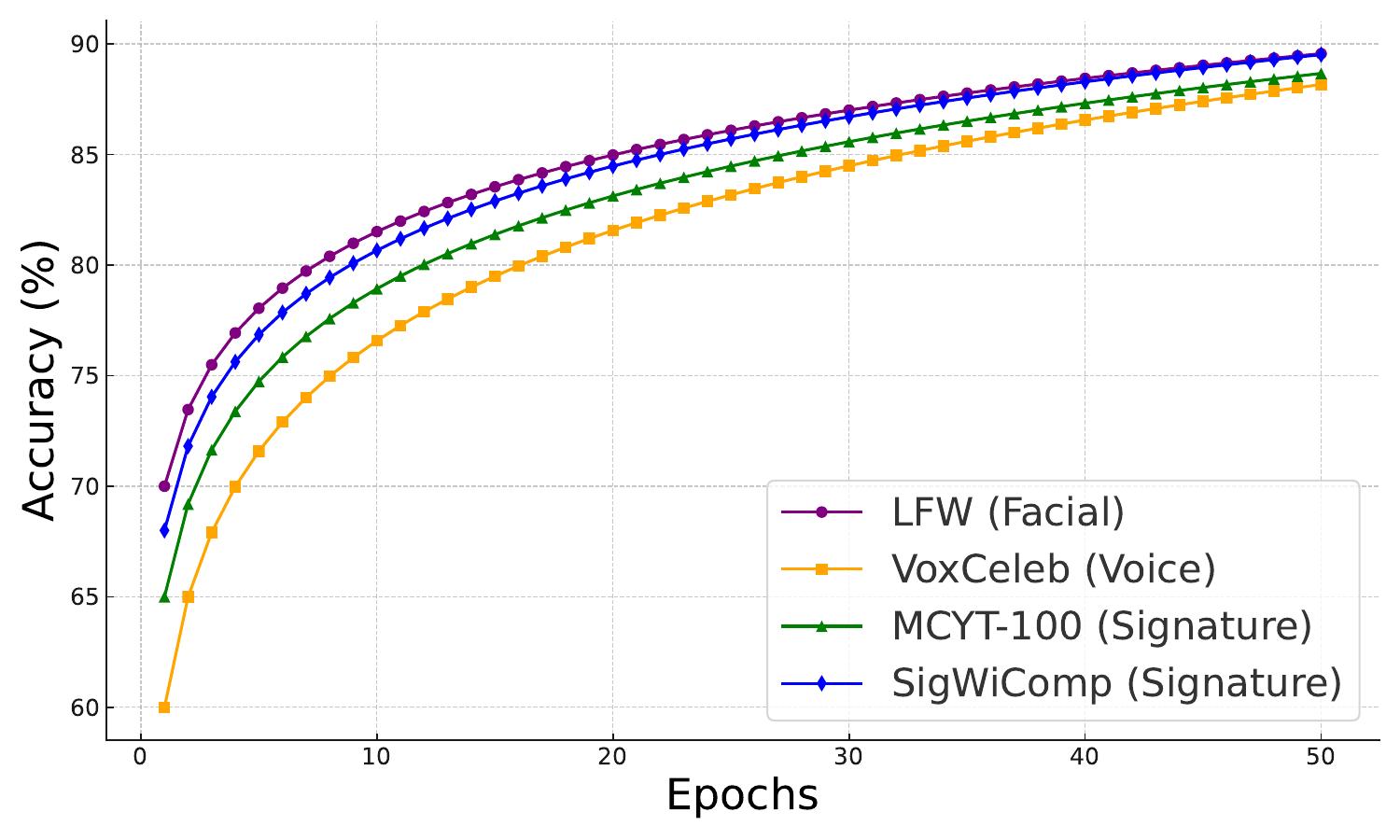}
    \caption{cross dataset testing}
    \label{fig6}
\end{figure}

\section{CONCLUSION}\label{sec5}
In conclusion, the multi-modal biometric authentication system, leveraging a shared-layer architecture, has been rigorously tested and validated through extensive simulations. The system exhibits a high accuracy rate, underpinned by a solid data collection and preprocessing strategy, as evidenced by the declining error rates across training epochs. With an AUC of 0.85, the system's ROC curve demonstrates its adeptness at balancing sensitivity and specificity, a direct result of the synergistic integration of spatial and temporal biometric data. The effective management of computational resources further reinforces the system's readiness for deployment in secure authentication scenarios, proving its capability to evolve with user feedback and adapt to emerging security challenges.  However, the system cannot fully ensure privacy as it requires cross-data learning between shared and modality-specific layers, which inherently involves shared access to biometric data from different modalities. Our future research will explore ways to mitigate this concern, potentially through more secure architecture designs or advanced privacy-preserving techniques that minimize data exposure while maintaining the model’s performance.

\bibliographystyle{plainnat}

\bibliography{ref}

\begin{IEEEbiography}[{\includegraphics[width=1in,height=1.25in,clip,keepaspectratio]{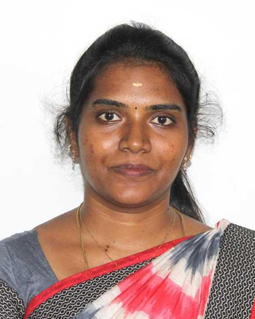}}]{Vatchala S} completed her Ph.D in the Department of Information and communication  Engineering, Anna University, Chennai. she is currently    working as an Assistant Professor senior grade-I at Vellore Institute of Technology, Chennai. She has published her research work in reputed Journals. She has also published 2 patents and one book under Lulu Press. Her current research interests include cloud security, Information security, Network Security \& Ethical Hacking.  
\end{IEEEbiography}

\begin{IEEEbiography}[{\includegraphics[width=1in,height=1.25in,clip,keepaspectratio]{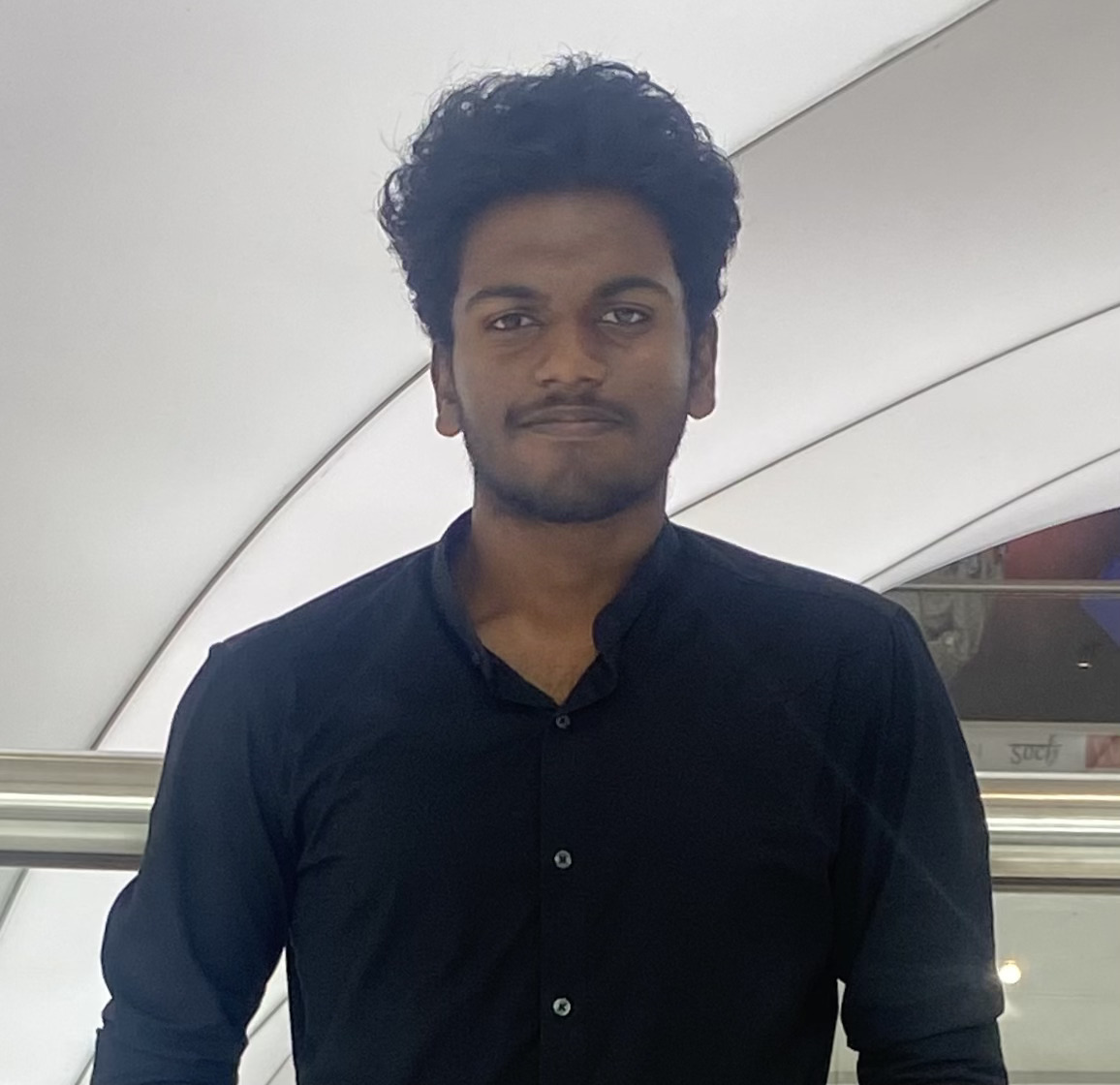}}]{Yeshwanth Govindarajan} is a current B.Tech student in Computer Science Engineering at VIT, Chennai, with research interests in artificial intelligence, ML, and neural networks. Currently interning and gaining practical experience at highperformr.ai, he actively contributes to the application of these technologies in real-world scenarios.
\end{IEEEbiography}

\begin{IEEEbiography}[{\includegraphics[width=1in,height=1.25in,clip,keepaspectratio]{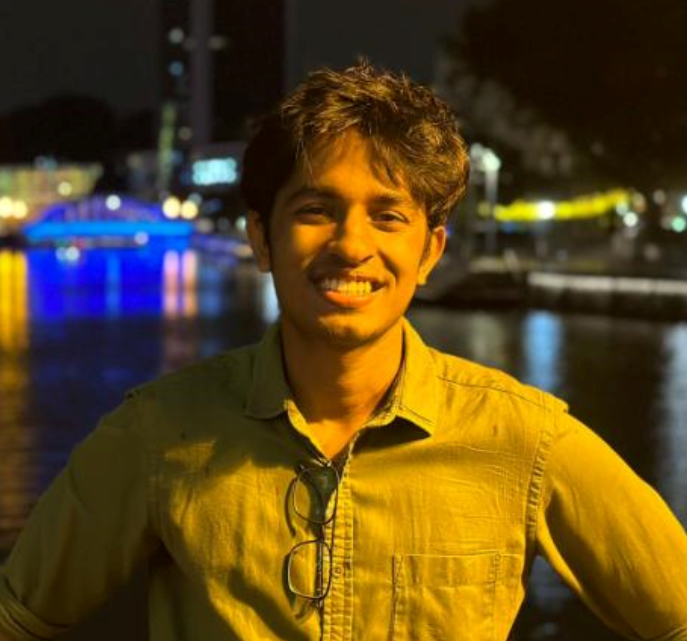}}]{Dharun Ramesh} is a B.Tech student in Computer Science Engineering at VIT, Chennai, focusing on artificial intelligence, machine learning, and neural networks. He is currently interning at Samsung R\&D Institute India, where he applies his knowledge to practical projects and real-world applications.
\end{IEEEbiography}

\begin{IEEEbiography}[{\includegraphics[width=1in,height=1.25in,clip,keepaspectratio]{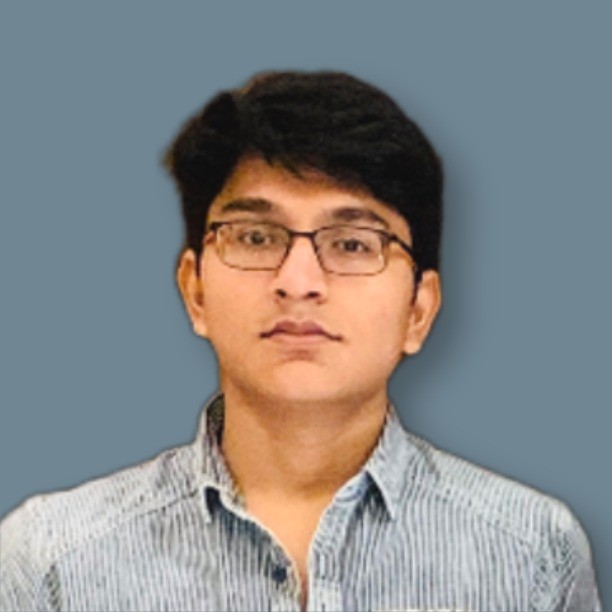}}]{Krithik Raja M} is a passionate researcher with a keen interest in ML models. As a full stack developer, he possesses the skills to create comprehensive solutions, addressing complex problems from end to end. His expertise in both front-end and back-end development enables him to seamlessly integrate advanced algorithms into practical applications, driving innovationandefficiency.
\end{IEEEbiography}

\begin{IEEEbiography}[{\includegraphics[width=1in,height=1.25in,clip,keepaspectratio]{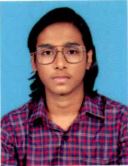}}]{Vishal Pranav Amirth Ganesan} is an aspiring B.Tech student in Computer Science Engineering at VIT University, with a keen interest in artificial intelligence, ML, and neural networks. Currently interning at highperformr.ai, he is passionate about applying these technologies to solve real-world challenges and actively contributing to innovative projects.
\end{IEEEbiography}

\begin{IEEEbiography}[{\includegraphics[width=1in,height=1.25in,clip,keepaspectratio]{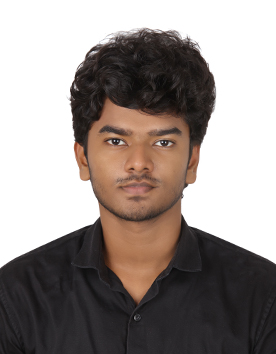}}] {Aashish Vinod Arul} is a dedicated B.Tech student in Computer Science \& Engineering at Vellore Institute of Technology, Chennai Campus, with a strong interest in algorithmic trading, ML, and data analytics. He is enthusiastic about leveraging these technologies to address real-world problems and is actively involved in pioneering projects such as advanced trading bots and AI-powered facial recognition systems..
\end{IEEEbiography}

\begin{IEEEbiography}[{\includegraphics[width=1in,height=1.25in,clip,keepaspectratio]{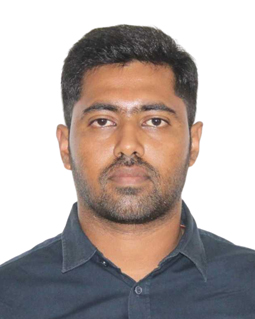}}] {Yogesh C} is currently an Associate Professor at School of Computing and Engineering, Vellore Institute of Technology, Chennai Campus. He obtained his PhD in Computer Science and Engineering at Universiti Malaysia Perlis, Malaysia. His area of research interests are ML, Feature Extraction and Selection, Optimization and Deep learning.He has published his research work in reputed Journals. he has also published 2 patents. He is an active researcher in many international journals.
\end{IEEEbiography}

\EOD

\end{document}